\documentclass[runningheads]{llncs}
\usepackage[T1]{fontenc}
\usepackage{graphicx}
\usepackage{algorithm}
\usepackage{algorithmic}
\usepackage{amsfonts}
\usepackage{amsmath}
\usepackage{booktabs}
\usepackage{bbm}
\usepackage{amsbsy}
\usepackage{multirow}
\usepackage{multicol}
\usepackage{makecell}
\usepackage{xcolor,colortbl}
\usepackage[colorlinks=true,
            linkcolor=blue,
            urlcolor=blue,
            citecolor=blue]{hyperref}

\usepackage{pifont}
\usepackage{xcolor}
\newcommand{\cmark}{\textcolor{green!80!black}{\ding{51}}}
\newcommand{\xmark}{\textcolor{red}{\ding{55}}}

\def\ourmodel{GTAD}
\def\ourmodels{GTAD }
\begin{document}
%
\title{Multi-Modal Graph Neural Network with Transformer-Guided Adaptive Diffusion for Preclinical Alzheimer Classification}

%
%

\author{Jaeyoon Sim\inst{1}
\and
Minjae Lee\inst{1}
\and
Guorong Wu\inst{2}
\and
Won Hwa Kim\inst{1}}
%
\authorrunning{J. Sim et al.}
\titlerunning{Multi-Modal GNN with Transformer-guided Adaptive Diffusion}


\institute{
Pohang University of Science and Technology, Pohang, South Korea \\
\email{\{simjy98, lalswo010, wonhwa\}@postech.ac.kr} \and
University of North Carolina at Chapel Hill, Chapel Hill, USA 
}

%
\maketitle              
\begin{abstract}
The graphical representation of the brain offers critical insights into diagnosing and prognosing neurodegenerative disease via relationships between  regions of interest (ROIs).
Despite recent emergence of various Graph Neural Networks (GNNs) to effectively capture the relational information, 
there remain inherent limitations in interpreting the brain networks.
Specifically, convolutional approaches ineffectively aggregate information from distant neighborhoods, 
while attention-based methods exhibit deficiencies in capturing node-centric information,
particularly in retaining critical characteristics from pivotal nodes. 
These shortcomings reveal challenges 
for identifying disease-specific variation from diverse features from different modalities.  
In this regard, we propose an integrated framework guiding diffusion process at each node by a downstream transformer 
where both short- and long-range properties of graphs are aggregated via diffusion-kernel and multi-head attention respectively. 
We demonstrate the superiority of our model 
by improving performance of pre-clinical Alzheimer’s disease (AD) classification 
with various modalities.
Also, our model adeptly identifies key ROIs that are closely associated with the preclinical stages of AD, 
marking a significant potential for early diagnosis and prevision of the disease.

\end{abstract}

\section{Introduction}
\label{sec:intro}


Amyloid deposition and neurofibrillary tangles disrupt neural connections, indicating the potential of using brain connectomes in neuroimaging 
to identify early signs of brain disorders such as Alzheimer’s Disease (AD) \cite{ad1,ad2,kim2015multi,ad3}.
The white-matter connectome establishes structural interconnections between distinct anatomical regions of interest (ROI) within the brain, 
and it constructs a brain network per subject. 
As the brain network guides pathological variation on the ROIs \cite{path_vari}, 
it is critical to incorporate the connection information in addition to regional measures from other images, e.g., magnetic resonance image (MRI) and positron emission tomography (PET) scans with various tracers, to characterize preclinical/early symptoms of AD. 





A typical representation of a brain network involves a graph, mathematically formulated by its nodes and edges. 
The nodes correspond to each ROI, and connectome features, e.g., number of tracts between ROIs and fractional anisotropy (FA), 
determine the edges with strength (i.e., edge weight). 
The graph representations of brain networks, 
together with image-derived measurements at each ROI, 
naturally lift the utilization of a graph neural network (GNN) for disease classification and characterization. 

Traditionally well-known GNNs \cite{gcn,gat} incorporate the structure of graphs via graph convolution, and 
later methods use kernel convolutions based on diffusion process to obtain better representations \cite{graphheat,gdc,adc,lsap}. 
These conventional methods rely on the homophily condition that node features locally connected by the edges 
behave similarly, 
overlooking the relationships between nodes far apart. 
Graph Transformers use global attention to capture far-distance influence beyond neighboring nodes
within the graph \cite{nodeformer,sgformer,difformer}, 
however, 
these methods often disregard sufficient expressive power of the central nodes,  
lacking interpretation of the result.  
The problem becomes more challenging when using multiple biomarker magnitudes as nodal features, as 
the interaction among multiple biomarkers and their diverse characteristics introduce heterogeneity, further complicating the analysis.


Therefore, it is necessary to develop an interpretable multi-modal method to 
capture both local characteristic and global graph-level information. 
The architecture we propose, i.e., {\bf G}NN with {\bf T}ransformer-guided {\bf A}daptive {\bf D}iffusion (\ourmodel), addresses the issues above by 
learning node-centric parameters of a diffusion kernel which are governed by a transformer. 
The encoder part of \ourmodel{} first obtains locally-effective representation of each node per imaging modality with a heat-kernel, 
which is later mixed by multi-head attention in the transformer to achieve globally-effective representation for classification. 
The node-wise kernel parameter as well as the attention scores let us 
interpret the local and global graph characteristics learnt by the model, especially 
when each node corresponds to anatomical ROI in the brain network. 


{\bf The key contributions of our work} are 
{\bf 1)} proposing a novel framework to aggregate both short- and long-range properties for better prediction of graph labels,
{\bf 2)} demonstrating superior performance on graph classification in comparison to the state-of-the-art methods, and
{\bf 3)} showing interpretability on the brain networks in a scenario with multiple imaging biomarkers.
Experiments on structural brain networks from Diffusion Tensor Imaging (DTI) and ROI measures from functional imaging from Alzheimer's Disease Neuroimaging Initiative (ADNI) study 
show that the developed framework 
yields practical results for pre-clincal AD classification and interpretation to facilitate early diagnosis and prevention of AD.

\begin{figure*}[!t]
\centering
    \scalebox{0.99}{
    \includegraphics[width=1.0\linewidth]{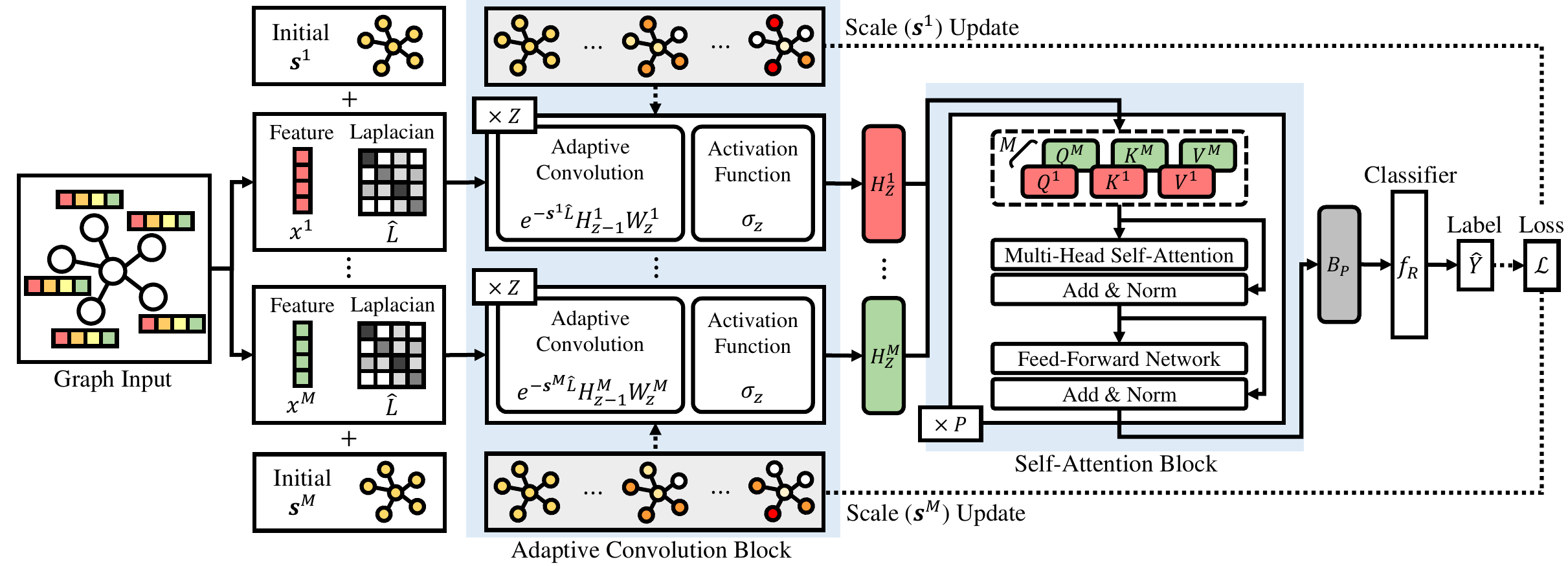}}
    \caption{
    Illustration of \ourmodel.
    A graph (as $\hat{L}$) and node feature $\textbf{\textit{x}}^m$ are inputted to $m$-th encoder at the adaptive convolution block.
    Then, all outputs $\{H^m_Z\}_{m=1}^M$ from this block are inputted to the self-attention block,
    producing an output $B_P$.
    Finally, the $B_P$ is entered into a classifier $f_R$ which yields a prediction $\hat{Y}$.
    To adaptively adjust the node-wise scales for each modality,
    the loss $\mathcal{L}$ from $\hat{Y}$ is backpropagated to update $m$-th encoder with scales $\textbf{\textit{s}}^m$.
    }
\label{fig:overall_architecture}
\end{figure*}

\section{Method}
\label{sec:method}




\noindent\textbf{Prelim: Graph Kernel Convolution.}
An undirected graph $G=\{V, E\}$ with $N$ nodes comprises a node set $V$ and an edge set $E$. 
A symmetric adjacency matrix $A$ and a diagonal degree matrix $D$ can be computed from $E$, whose elements encode
connectivity among its nodes and 
the volume of each node respectively. 
A graph Laplacian is defined as $L=D-A$,  
which is real and positive semi-definite. 
It has a complete set of orthonormal eigenvectors $U = [u_{1} | u_{2} | ... | u_{N}]$ 
and corresponding real and non-negative eigenvalues $0=\lambda_{1} \leq \lambda_{2} \leq ... \leq \lambda_{N}$, so does the normalized Laplacian 
$\hat{L}=D^{-1/2}LD^{-1/2}$. 

From Spectral Graph Theory \cite{spectral}, the choice of a kernel function determines specific graph characteristics.
For example, a prominent heat-kernel 
between nodes $p$ and $q$ 
is spanned by $U$ as 
\begin{equation}
    h_s(p, q) = \sum_{i=1}^{N}e^{-s\lambda_{i}}u_{i}(p)u_{i}(q)
    \label{eq:heat_kernel_between_nodes}
\end{equation}
where $u_i$ is the $i$-th eigenvector. 
The kernel $e^{-s\lambda_i}$ captures smooth transition between the nodes 
within the scale $s$ as a low-pass filter. 
Using convolutional theorem \cite{signals}, 
graph Fourier transform, i.e., $\hat{x}=U^Tx$, 
defines the graph convolution $\ast$ of a signal $x(p)$ with a filter $h_s$ as
\begin{equation}
    h_s \ast x(p) = \sum_{i=1}^{N}e^{-s\lambda_{i}}\hat{x}(i)u_{i}(p)
    \label{eq:heat_kernel_convolution}
\end{equation}
whose band-width is controlled by the scale $s$.

\noindent\textbf{Modality-wise Adaptive Convolution Block.}
Consider a graph $G$ given as a normalized Laplacian $\hat L \in \mathbb{R}^{N\times N}$, 
a set of features (i.e., imaging measures) 
$X = \{\textbf{\textit{x}}^m\}_{m=1}^M$ defined on $N$ nodes from $M$ modalities, a set of trainable multi-variate scales $\{\textbf{\textit{s}}^m\}_{m=1}^M$ where $\textbf{\textit{s}}^m \in \mathbb R^N$ 
and a graph label $Y$.
To obtain representations from individual modality, 
encoders from our model take $\hat L$ and $\textbf{\textit{x}}^m$ for $m\in\{1,\dots,M\}$ as inputs 
and perform convolution with heat kernel in Eq.~\eqref{eq:heat_kernel_convolution} across all nodes and modalities respectively.
Each encoder consists of multiple graph convolution layers that adaptively aggregate features for each node with a non-linear activation function $\sigma_z$ as
\begin{equation}
    H_z^m = \sigma_z(e^{-\textbf{\textit{s}}^m\hat L}H_{z-1}^mW_z^m)
    \label{eq:mm_graph_convolution}
\end{equation}
where $H_z^m$ is an output from $z$-th convolution layer for $m$-th modality with $H_0^m=\textbf{\textit{x}}^m$, 
and $W_z^m$ is a weight matrix. 
Within our framework, $\textbf{\textit{s}}^m$ is made trainable to capture local 
characteristic of individual node for different modalities.
Since the Eq.~\eqref{eq:mm_graph_convolution} is an operation in a single convolution layer, 
better representation for the original feature $\textbf{\textit{x}}^m$ can be achieved by stacking $Z$ of them.

\noindent\textbf{Modality-wise Self-Attention Block.}
The obtained embeddings $\{H_Z^m\}_{m=1}^M$ are inputted to an attention block to compute node-wise attention scores.
Here, the multi-head self-attention module is inherited from the transformer layer~\cite{transformer}.
Unlike typical use of transformer~\cite{bert,worth},
each head is assigned to an individual modality 
to integrate {\em long-range} information from other nodes, 
which is not captured in the convolution block.

The input of attention module consists of query $Q^m \in \mathbb{R}^{N\times C}$, key $K^m \in \mathbb{R}^{N\times C}$, and value $V^m \in \mathbb{R}^{N\times C}$ from modality-wise embedding $H_Z^m$,
and $C$ is the dimension for hidden units.
The self-attention scores are computed as $Q^m{K^m}^T / \sqrt C$, 
and softmax $\sigma$ is applied to obtain weights on the values.
Using the self-attention scores, 
a self-attention value is computed as
\begin{equation}
    \phi(Q^m,K^m,V^m) = \sigma(\frac{Q^m{K^m}^T}{\sqrt C})V^m.
    \label{lab:attention_values}
\end{equation}

As a single attention head is assigned to a single modality,
the global characteristics for all modalities is averaged with 
a multi attention function as $\Phi(Q,K,V) = [h^1|h^2|\dots|h^m]W^\Phi$
where $h^m = \phi(Q^mW^{Q^m},K^mW^{K^m},V^mW^{V^m})$. 
Here, $W^{Q^m}$, $W^{K^m}$, $W^{V^m}$ and $W^\Phi$ are weight matrices for $Q^m$, $V^m$, $K^m$ and $\Phi(\cdot)$ respectively.
Thus, multi-modal self-attention enables the model to jointly attend to information from different modalities across various ROIs in long-range.

The attention block contains a fully connected feed-forward module $\Psi(\cdot)$,
which consists of multiple linear transformations with a non-linear activation function in between.
To stabilize learning process and improve generalization,
residual connections~\cite{residual} are used,
followed by layer normalization $f_{L}[\cdot]$~\cite{layernorm}.
Therefore, a comprehensive context across all nodes is captured as 
\begin{equation}
    B_p=f_{L}[f_{L}[B_{p-1} + \Phi(B_{p-1})] +
    \Psi(f_{L}[B_{p-1} + \Phi(B_{p-1})])]
    \label{eq:mhsa}
\end{equation}
where $B_p$ is an output from $p$-th attention layer,
and multi-modal representations $\{H_Z^m\}_{m=1}^M$ are used as $Q$, $K$ and $V$ for $B_0$.
To capture complex dependencies in the input modalities, 
multiple attention layers, e.g., $P$-layers, can be stacked.

\noindent\textbf{Transformer-Guided Scale Update.} 
Consider a set of graphs $\{G_t\}_{t=1}^T$ with corresponding labels $\{Y_t\}_{t=1}^T$,
and learning a classification model finds a function $f(G_t)=Y_t$. 
For this, a downstream classifier $f_R(\cdot)$ 
takes the $B_P$ from Transformer as an input and returns a prediction $\hat Y_{tj}$ 
at the $j$-th class for the $t$-th sample, which is computed via $Softmax$ as 
\begin{equation}
    \hat Y_{tj} = \frac{f_R(B_P)_{tj}}{\sum_{j^{'}\in J} f_R(B_P)_{tj^{'}}}
    \label{eq:prediction}
\end{equation}
where $J$ is a class size. 
To update a scale $s_{n}^m$ at the $n$-th node for the $m$-th encoder,
the objective function is defined by cross-entropy between the true value $Y_{tj}$ 
and the prediction $\hat Y_{tj}$. 
With an $\ell_1$ norm regularization on $s_{n}^m$ to impose positive scale for the heat-kernel, 
the overall objective function $\mathcal{L}$ 
is defined as
\begin{equation}
    \mathcal{L} = -\frac{1}{T}\sum_{t=1}^T\sum_{j=1}^JY_{tj}\text{ln}\hat{Y}_{tj}+\alpha\frac{1}{M}\sum_{m=1}^M\sum_{n=1}^N \mathbbm{1}_{s<0}|s_n^m|
    \label{eq:loss_total}
\end{equation}
where $\alpha$ is a user-parameter and $\mathbbm{1}$ is an indicator function. 
Update of the modality-specific scales 
is performed as $s\leftarrow s- \beta\frac{\partial \mathcal{L}}{\partial s}$ 
via gradient-descent with a learning rate $\beta$.

\section{Experiments}
\label{sec:experiments}

\textbf{Dataset.} 
Neuroimages of $T$=919 preclinical AD subjects 
in the Alzheimer's Disease Neuroimaging Initiative (ADNI) study were used for the experiment. 
Each brain was partitioned into 148 cortical regions and 12 sub-cortical regions with Destrieux atlas~\cite{destrieux} with MRI, and 
Tractography on diffusion weighted imaging (DWI) was applied to calculate the number of white matter fibers connecting the 160 brain regions to construct brain network. 
On the same parcellation, 
region-wise imaging features such as Standard Uptake Value Ratio (SUVR)~\cite{suvr} of metabolic intensity from FDG-PET, 
$\beta$-Amyloid protein from Amyloid-PET and cortical thickness from MRI were measured.
Each subject was assigned to Control (CN, $T$=333), Significant Memory Concern (SMC, $T$=172) and Early Mild Cognitive Impairment (EMCI, $T$=414) for group comparisons. 


\noindent\textbf{Setup.} We designed various 3-way classifications to classify the pre-clinical groups using various combinations of biomarkers.
5-fold cross validation was used to obtain unbiased results, 
and accuracy, precision and recall in their mean were computed for evaluation.
As the baselines, we categorized GNNs into three groups and adopted them; 
1) Convolution-based GNNs such as GCN \cite{gcn} and GAT \cite{gat}, 
2) GNNs with graph diffusion such as GraphHeat \cite{graphheat}, GDC \cite{gdc}, ADC \cite{adc} and LSAP \cite{lsap},
and 3) Graph transformers such as NodeFormer \cite{nodeformer}, DIFFormer \cite{difformer} and SGFormer \cite{sgformer}.
More details
are given in the supplementary.

\begin{table*}[!t]
\caption{
Preclinical AD classification performance (CN/SMC/EMCI) on ADNI data.
}
\centering
\renewcommand{\arraystretch}{1.0}
\renewcommand{\tabcolsep}{0.23cm}
\scalebox{0.73}{
    \begin{tabular}{l||ccc|ccc}
    \Xhline{3\arrayrulewidth}
    \textbf{Modalities} & \multicolumn{3}{c|}{\textbf{Cortical Thickness \& $\beta$-Amyloid}} & \multicolumn{3}{c}{\textbf{Cortical Thickness \& FDG}} \\
    \hline
    \textbf{Methods} & \textbf{Accuracy} & \textbf{Precision} &  \textbf{Recall} & \textbf{Accuracy} & \textbf{Precision} &  \textbf{Recall} \\
    \hline
    GCN \cite{gcn} & 
    0.861$\pm$0.04 & 0.772$\pm$0.06  & 0.780$\pm$0.06 &  
    0.873$\pm$0.02 & 0.802$\pm$0.02  & 0.813$\pm$0.03 \\

    GAT \cite{gat} & 
    0.896$\pm$0.01 & 0.827$\pm$0.03 & 0.839$\pm$0.02 & 
    0.882$\pm$0.02 & 0.811$\pm$0.03 & 0.844$\pm$0.03 \\

    GraphHeat \cite{graphheat} & 
    0.868$\pm$0.02 & 0.777$\pm$0.05 & 0.797$\pm$0.04 &
    0.887$\pm$0.03 & 0.821$\pm$0.04 & 0.834$\pm$0.03 \\

    GDC \cite{gdc} & 
    0.858$\pm$0.02 & 0.767$\pm$0.03 & 0.786$\pm$0.04 & 
    0.842$\pm$0.01 & 0.743$\pm$0.02 & 0.765$\pm$0.03 \\

    ADC \cite{adc} & 
    0.906$\pm$0.02 & 0.835$\pm$0.03 & 0.861$\pm$0.04 & 
    0.896$\pm$0.01 & 0.831$\pm$0.01 & 0.847$\pm$0.02 \\


    LSAP \cite{lsap} & 
    0.911$\pm$0.01 & 0.847$\pm$0.03 & 0.872$\pm$0.02 & 
    0.934$\pm$0.02 & 0.899$\pm$0.05 & 0.904$\pm$0.03 \\

    NodeFormer \cite{nodeformer} & 
    0.916$\pm$0.02 & 0.856$\pm$0.04 & 0.865$\pm$0.02 & 
    0.944$\pm$0.01 & 0.913$\pm$0.03 & 0.921$\pm$0.02 \\

    DIFFormer \cite{difformer} & 
    0.930$\pm$0.01 & 0.877$\pm$0.03 & 0.900$\pm$0.02 & 
    0.954$\pm$0.01 & 0.923$\pm$0.02 & 0.944$\pm$0.01 \\

    SGFormer \cite{sgformer} & 
    0.941$\pm$0.01 & 0.894$\pm$0.03 & 0.911$\pm$0.02 & 
    0.959$\pm$0.01 & 0.931$\pm$0.01 & 0.945$\pm$0.01 \\

    \cellcolor{gray!20}\ourmodels(Ours) & 
    \cellcolor{gray!20}\textbf{0.945$\pm$0.02} & \cellcolor{gray!20}\textbf{0.901$\pm$0.03} & \cellcolor{gray!20}\textbf{0.919$\pm$0.02} & 
    \cellcolor{gray!20}\textbf{0.963$\pm$0.01} & \cellcolor{gray!20}\textbf{0.935$\pm$0.02} &  \cellcolor{gray!20}\textbf{0.948$\pm$0.01} \\
    
    \hline
    \textbf{Modalities} & \multicolumn{3}{c|}{\textbf{$\beta$-Amyloid \& FDG}} & 
    \multicolumn{3}{c}{\textbf{All Imaging Features}}
    \\
    \hline
    \textbf{Methods} & \textbf{Accuracy} & \textbf{Precision} & \textbf{Recall} & \textbf{Accuracy} & \textbf{Precision} &  \textbf{Recall} \\
    \hline
    
    GCN \cite{gcn} & 
    0.880$\pm$0.01 & 0.806$\pm$0.02 & 0.813$\pm$0.02 & 
    0.888$\pm$0.02 & 0.816$\pm$0.02 & 0.826$\pm$0.02 \\
    
    GAT \cite{gat} & 
    0.877$\pm$0.02 & 0.815$\pm$0.03 & 0.814$\pm$0.04 &
    0.912$\pm$0.01 & 0.858$\pm$0.02 & 0.864$\pm$0.02 \\

    GraphHeat \cite{graphheat} & 
    0.880$\pm$0.02 & 0.804$\pm$0.05 & 0.824$\pm$0.03 &
    0.893$\pm$0.02 & 0.824$\pm$0.03 & 0.839$\pm$0.03 \\

    GDC \cite{gdc} & 
    0.866$\pm$0.02 & 0.787$\pm$0.03 & 0.790$\pm$0.03 &
    0.867$\pm$0.02 & 0.779$\pm$0.03 & 0.799$\pm$0.02 \\

    ADC~\cite{adc} & 
    0.910$\pm$0.01 & 0.865$\pm$0.02 & 0.856$\pm$0.02 & 
    0.904$\pm$0.02 & 0.855$\pm$0.04 & 0.858$\pm$0.02 \\


    LSAP \cite{lsap} & 
    0.922$\pm$0.02 & 0.862$\pm$0.05 & 0.893$\pm$0.03 &
    0.912$\pm$0.01 & 0.844$\pm$0.04 & 0.879$\pm$0.02 \\

    NodeFormer \cite{nodeformer} & 
    0.931$\pm$0.01 & 0.887$\pm$0.03 & 0.893$\pm$0.03 & 
    0.938$\pm$0.02 & 0.900$\pm$0.03 & 0.902$\pm$0.03 \\

    DIFFormer \cite{difformer} & 
    0.951$\pm$0.01 & 0.919$\pm$0.03 & 0.933$\pm$0.02 &
    0.953$\pm$0.01 & 0.920$\pm$0.02 & 0.936$\pm$0.02 \\

    SGFormer \cite{sgformer} & 
    0.954$\pm$0.01 & 0.923$\pm$0.03 & 0.936$\pm$0.02 &
    0.951$\pm$0.01 & 0.911$\pm$0.02 & 0.933$\pm$0.02 \\
    
    \cellcolor{gray!20}\ourmodels(Ours) & 
    \cellcolor{gray!20}\textbf{0.962$\pm$0.01} & 
    \cellcolor{gray!20}\textbf{0.935$\pm$0.02} & 
    \cellcolor{gray!20}\textbf{0.946$\pm$0.02} & 
    \cellcolor{gray!20}\textbf{0.963$\pm$0.01} & 
    \cellcolor{gray!20}\textbf{0.943$\pm$0.01} & 
    \cellcolor{gray!20}\textbf{0.941$\pm$0.02} 
    \\
    
    \Xhline{3\arrayrulewidth}
    \end{tabular}
    }
\label{tab:performance}
\end{table*}

\noindent\textbf{Classification Result.}
The performance comparisons between our model and nine baselines across four experiments are reported in Table \ref{tab:performance}.
As shown in Table~\ref{tab:performance}, aggregating both local (i.e., short-range) features  by adaptively learned modality-wise scales and global (i.e., long-range) information by global attentions performed the best in all experimental cases,
and accuracy from most experiments showed over 96\% except for the case using cortical thickness and $\beta$-Amyloid.
Notably, \ourmodels outperformed the outstanding transformers in pre-clinical AD prediction, 
indicating that our model is more suitable on the brain network 
even under difficult conditions (i.e., prediction for early stages in AD given multiple imaging scans).
Also, the stability of our model can be explained by low standard deviations for all evaluations within 5-folds.

\section{Interpretation of the Trained \ourmodels}
\label{sec:analyses}

\begin{figure*}[!t]
\centering

    \renewcommand{\arraystretch}{0.6}
    \renewcommand{\tabcolsep}{0.01cm}
    \scalebox{0.9}{\scriptsize
    \begin{tabular}{cccccccl}
    & \multicolumn{2}{c}{\raisebox{0\height}[0pt][0pt]{\textbf{Cortical Thickness}}}
    & \multicolumn{2}{c}{\raisebox{0\height}[0pt][0pt]{\textbf{$\beta$-Amyloid}}}
    & \multicolumn{2}{c}{\raisebox{0\height}[0pt][0pt]{\textbf{FDG}}}
    & \\ 

    \quad\quad
    & \includegraphics[width=0.17\linewidth]{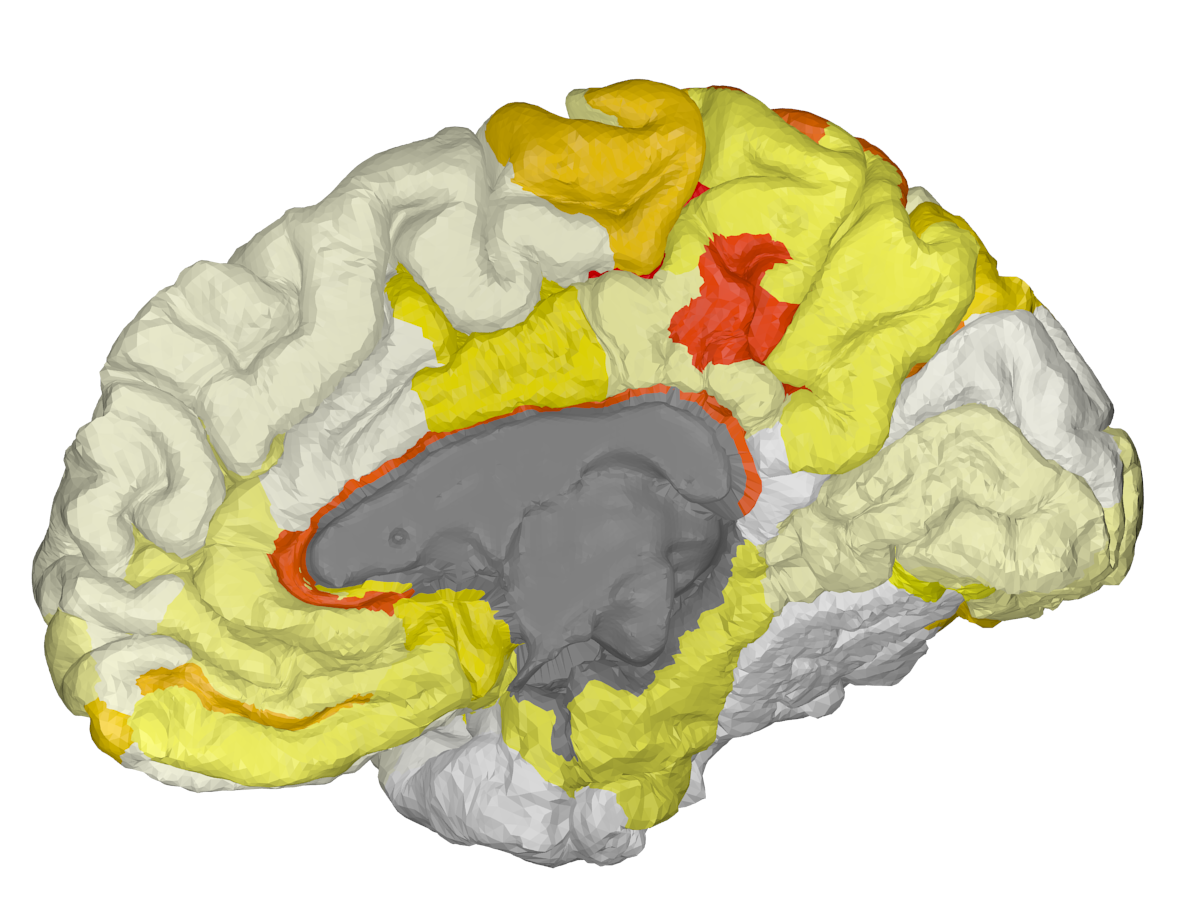}
    & \includegraphics[width=0.17\linewidth]{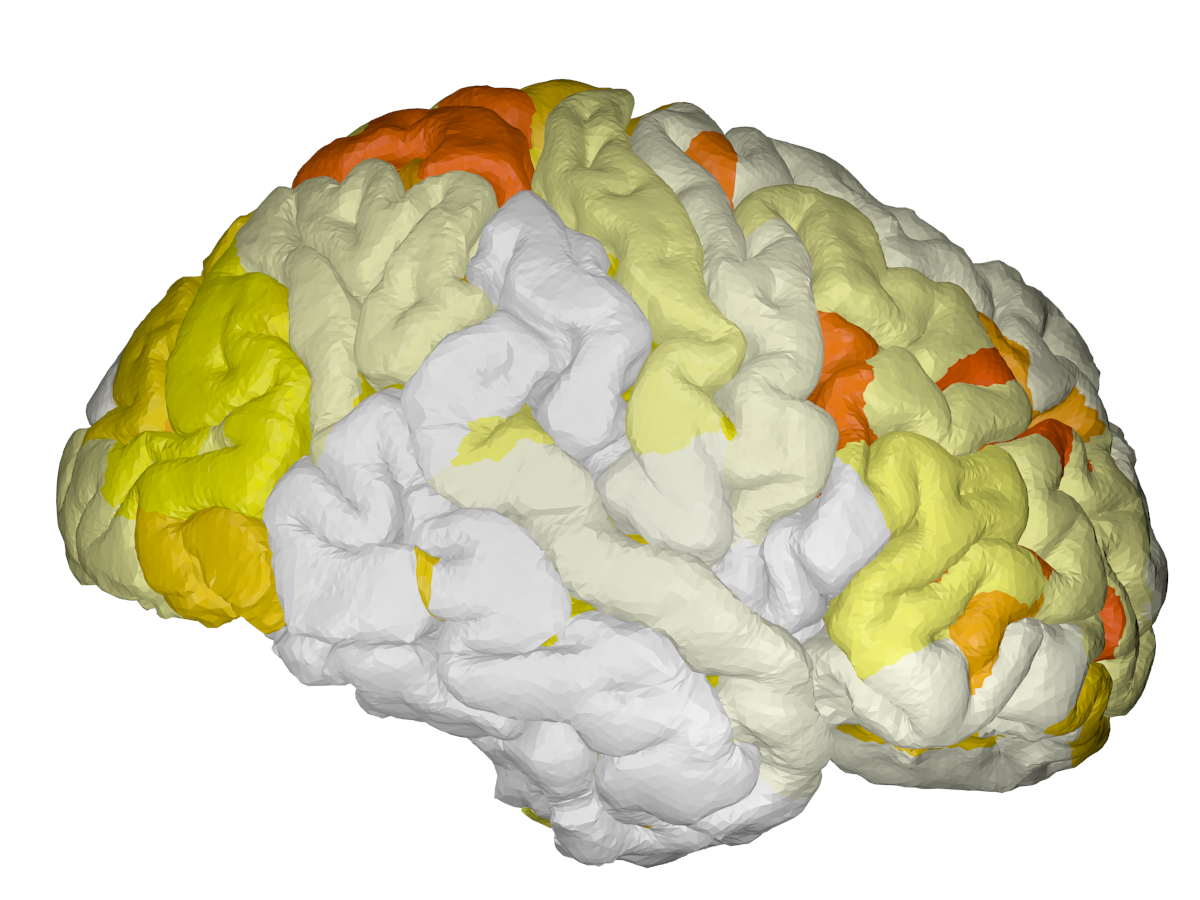}
    & \includegraphics[width=0.17\linewidth]{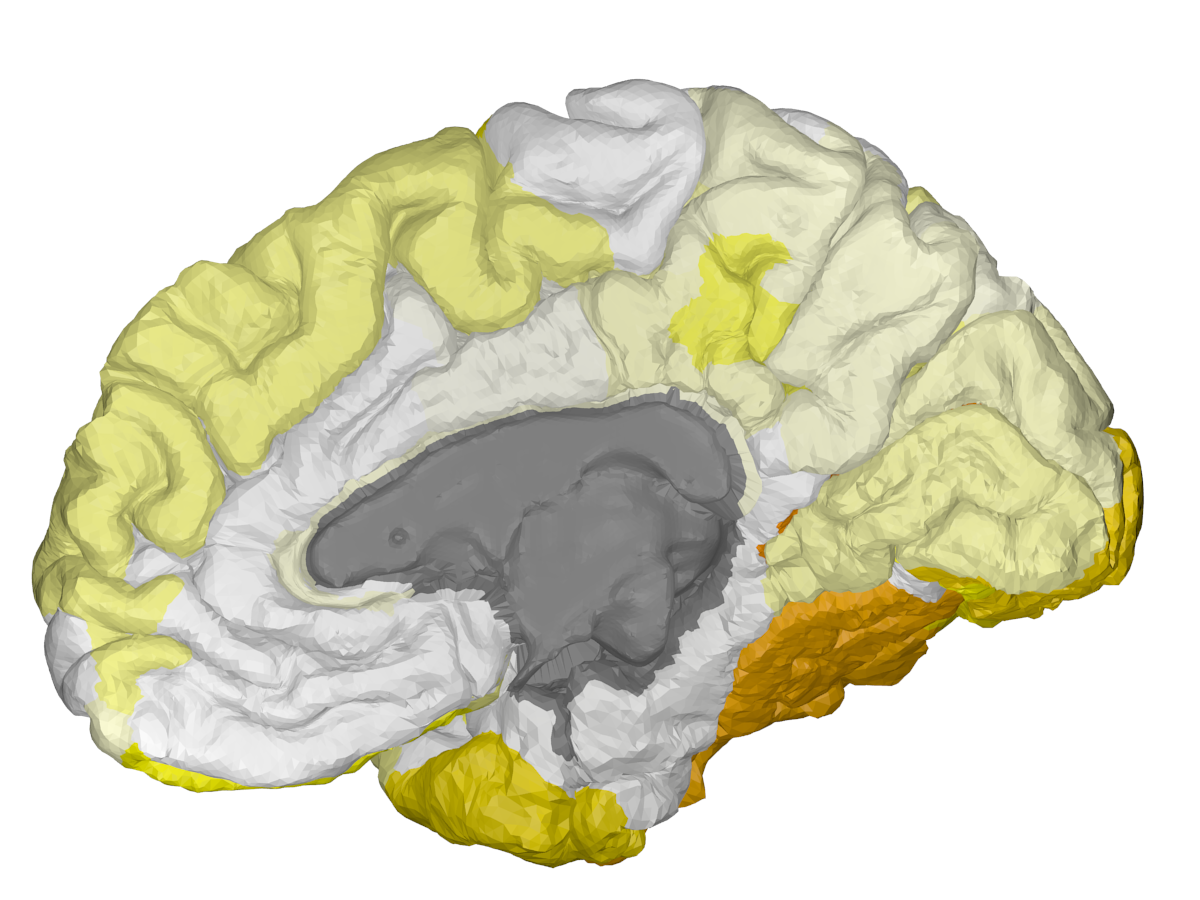}
    & \includegraphics[width=0.17\linewidth]{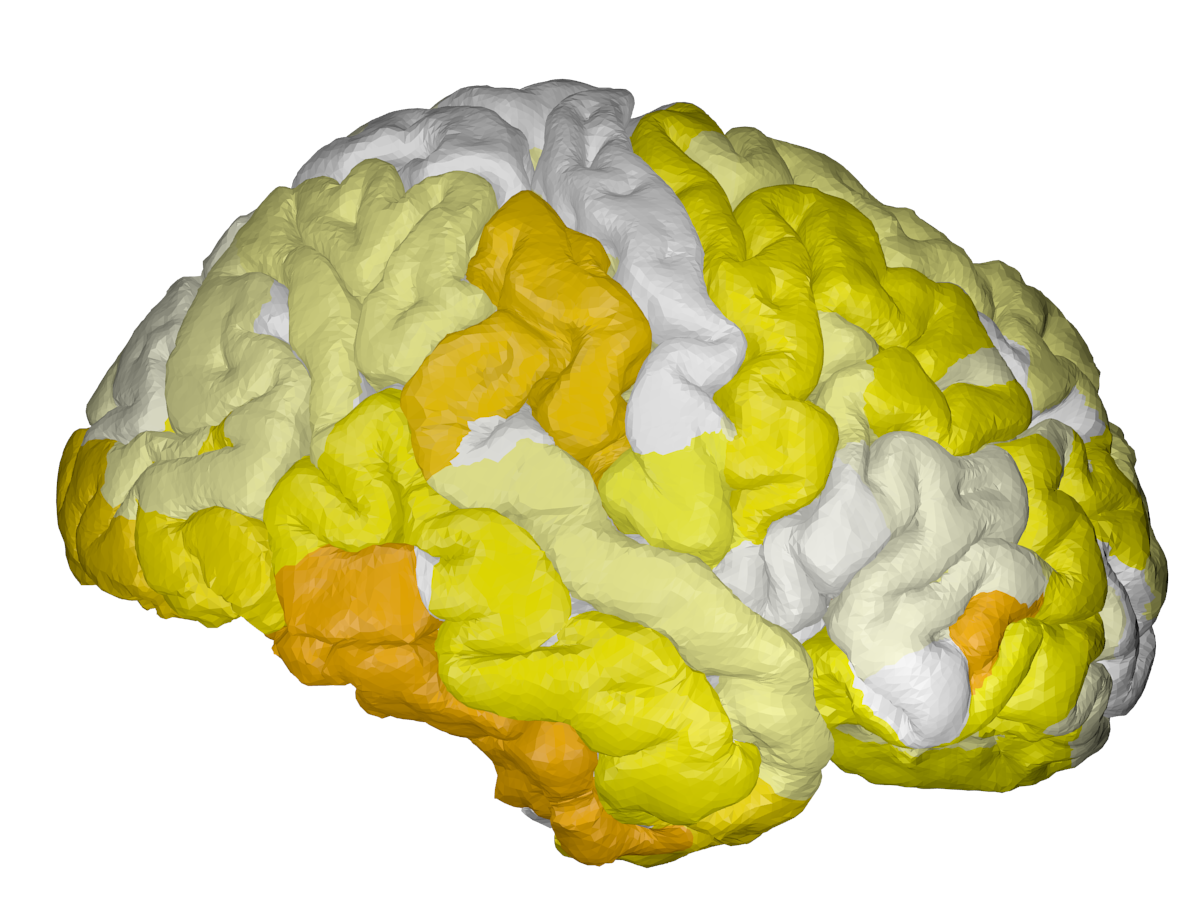}
    & \includegraphics[width=0.17\linewidth]{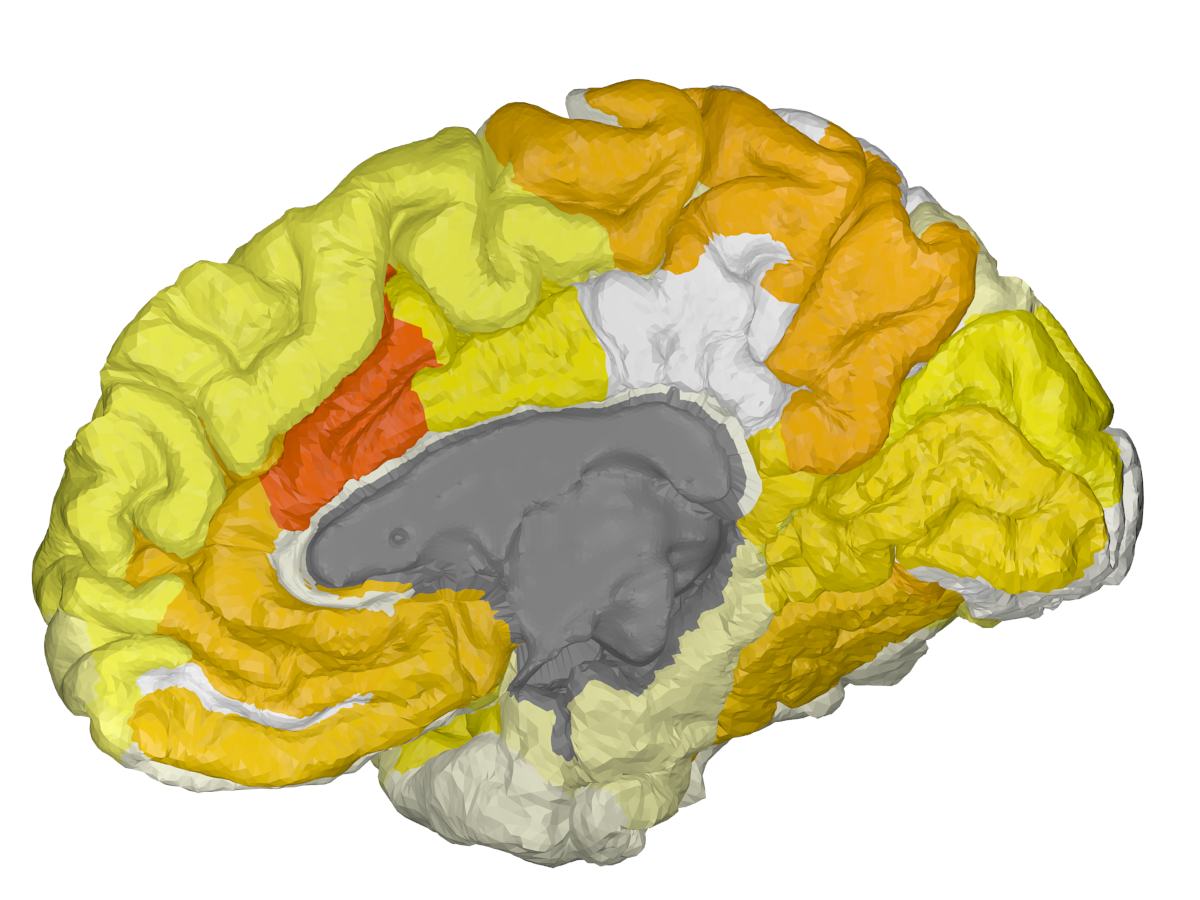}
    & \includegraphics[width=0.17\linewidth]{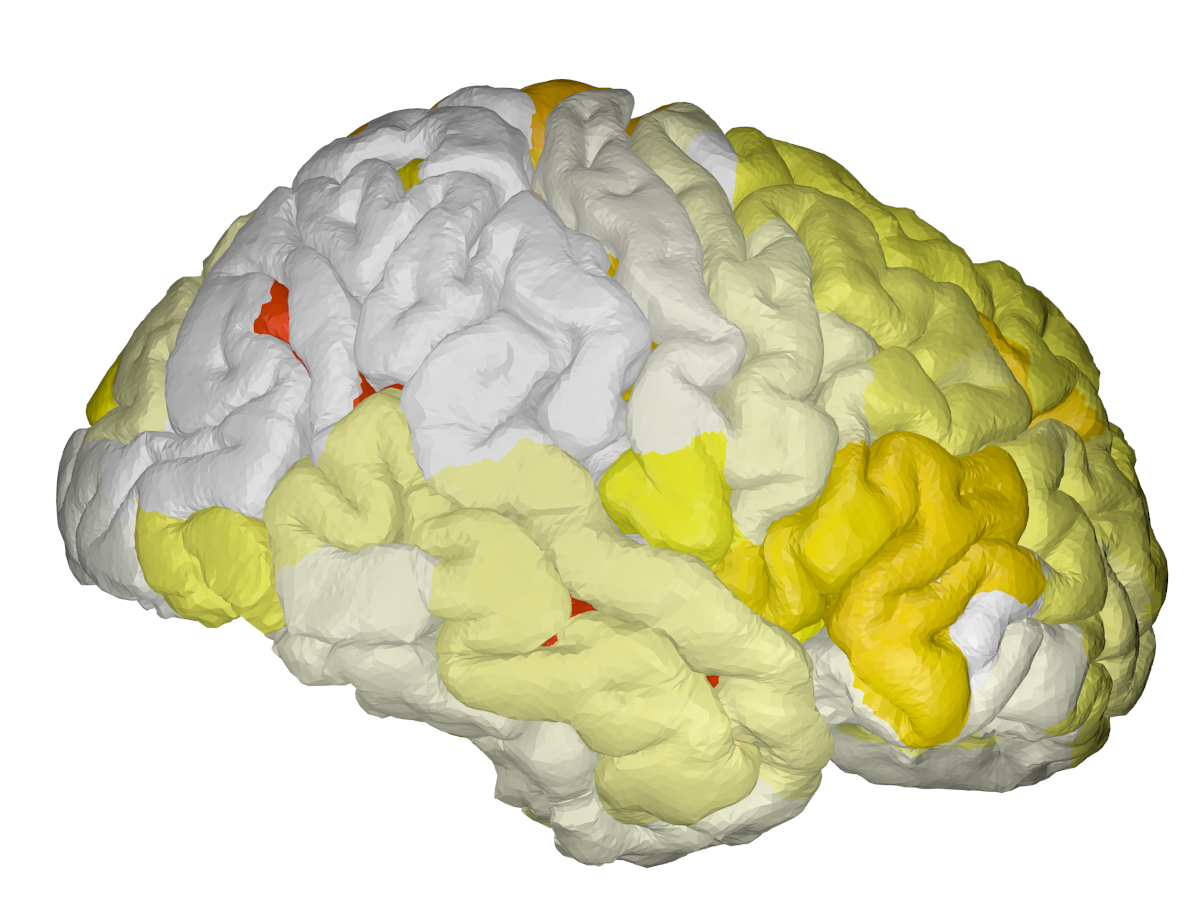}
    &\\
    
    \quad\quad
    & \includegraphics[width=0.17\linewidth]{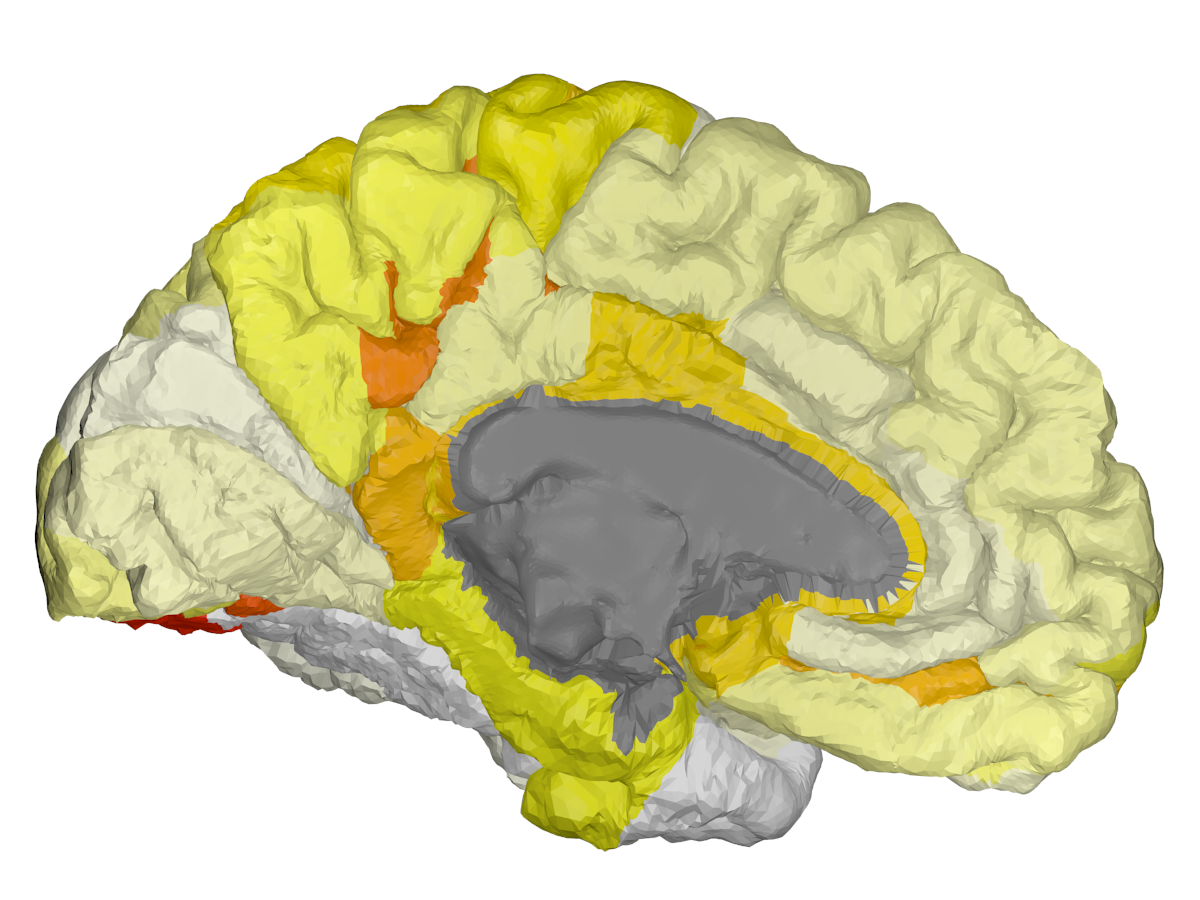}
    & \includegraphics[width=0.17\linewidth]{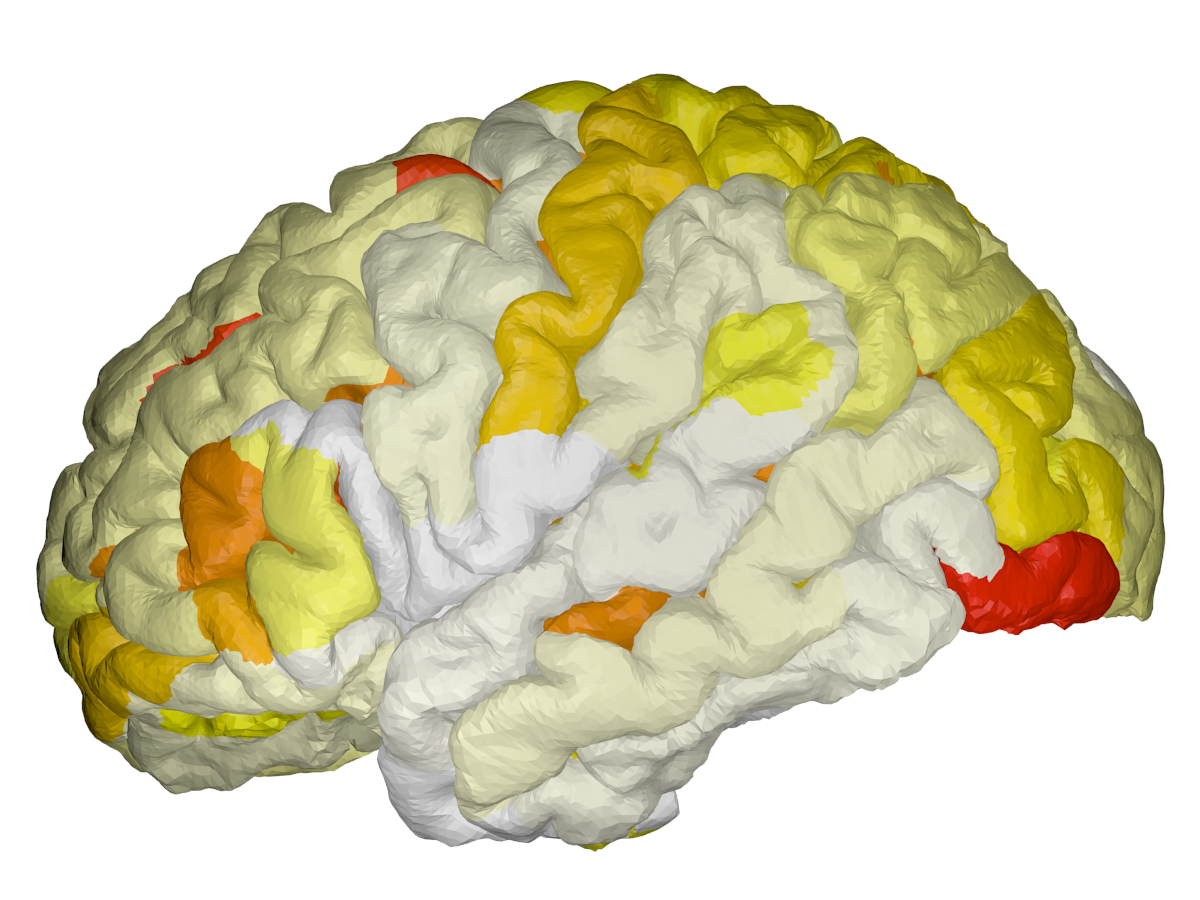}
    & \includegraphics[width=0.17\linewidth]{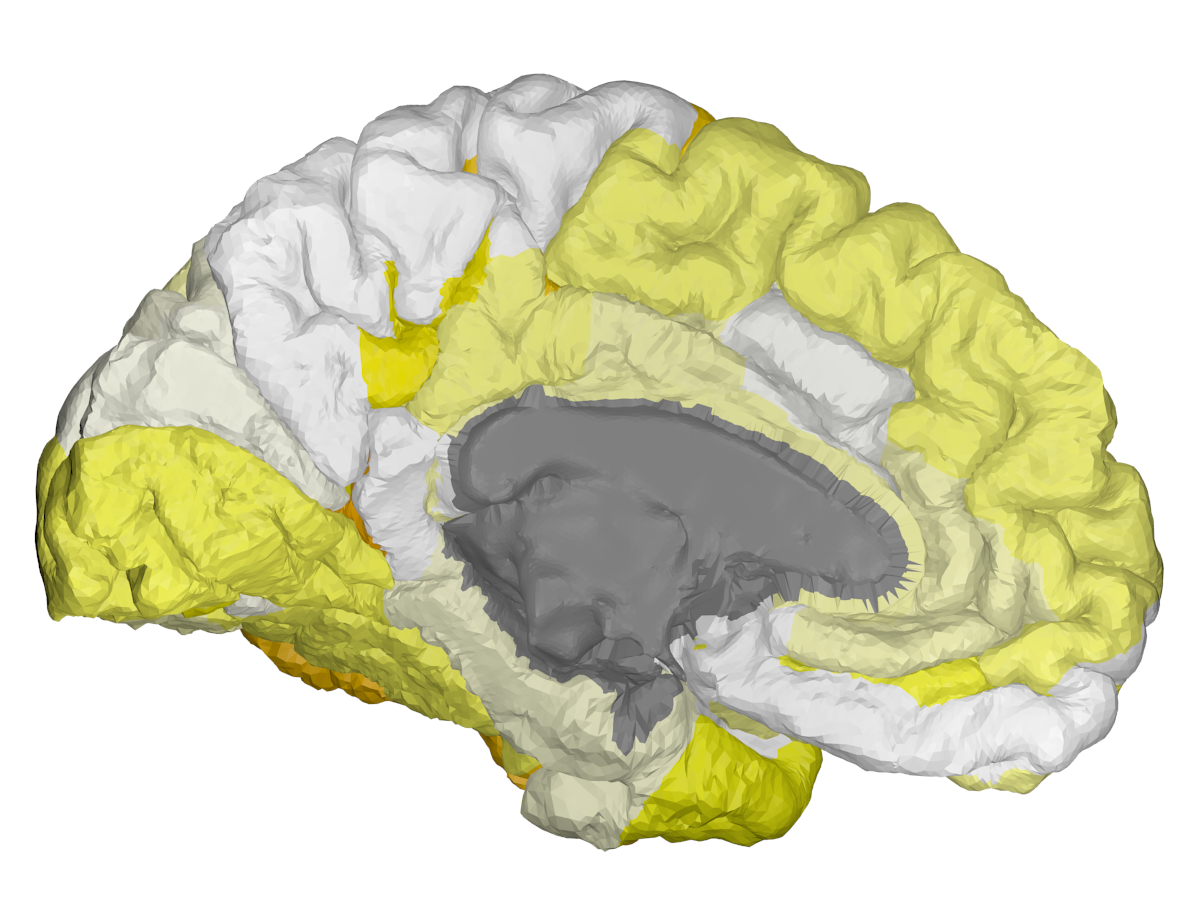}
    & \includegraphics[width=0.17\linewidth]{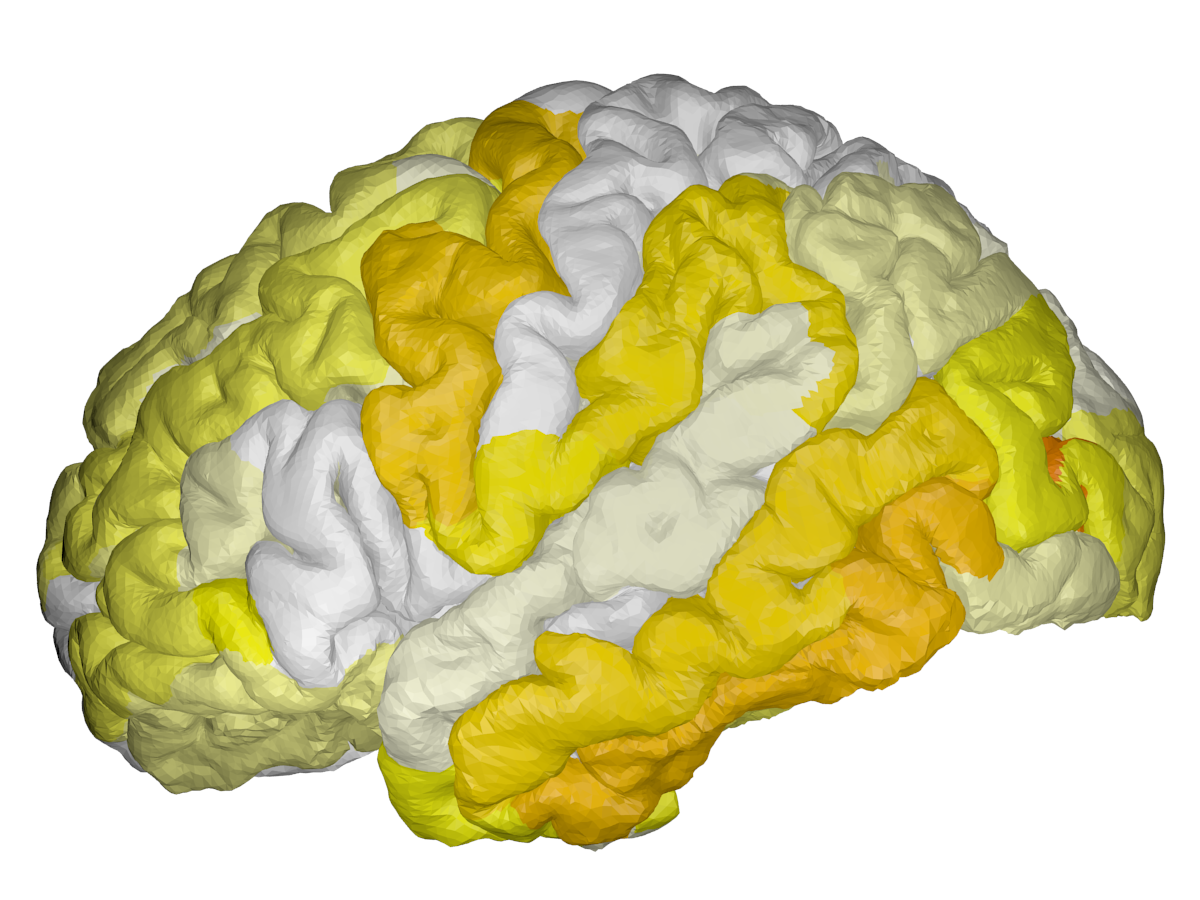}
    & \includegraphics[width=0.17\linewidth]{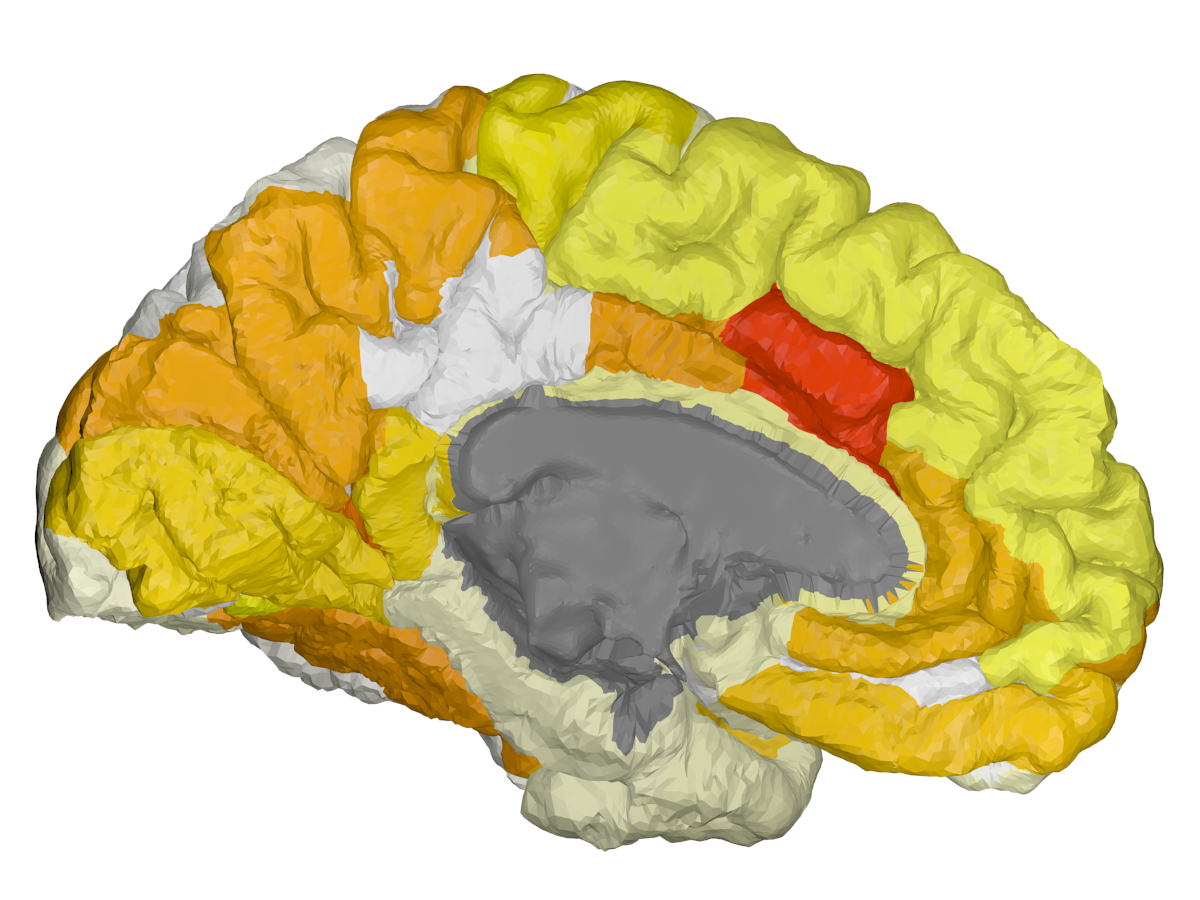}
    & \includegraphics[width=0.17\linewidth]{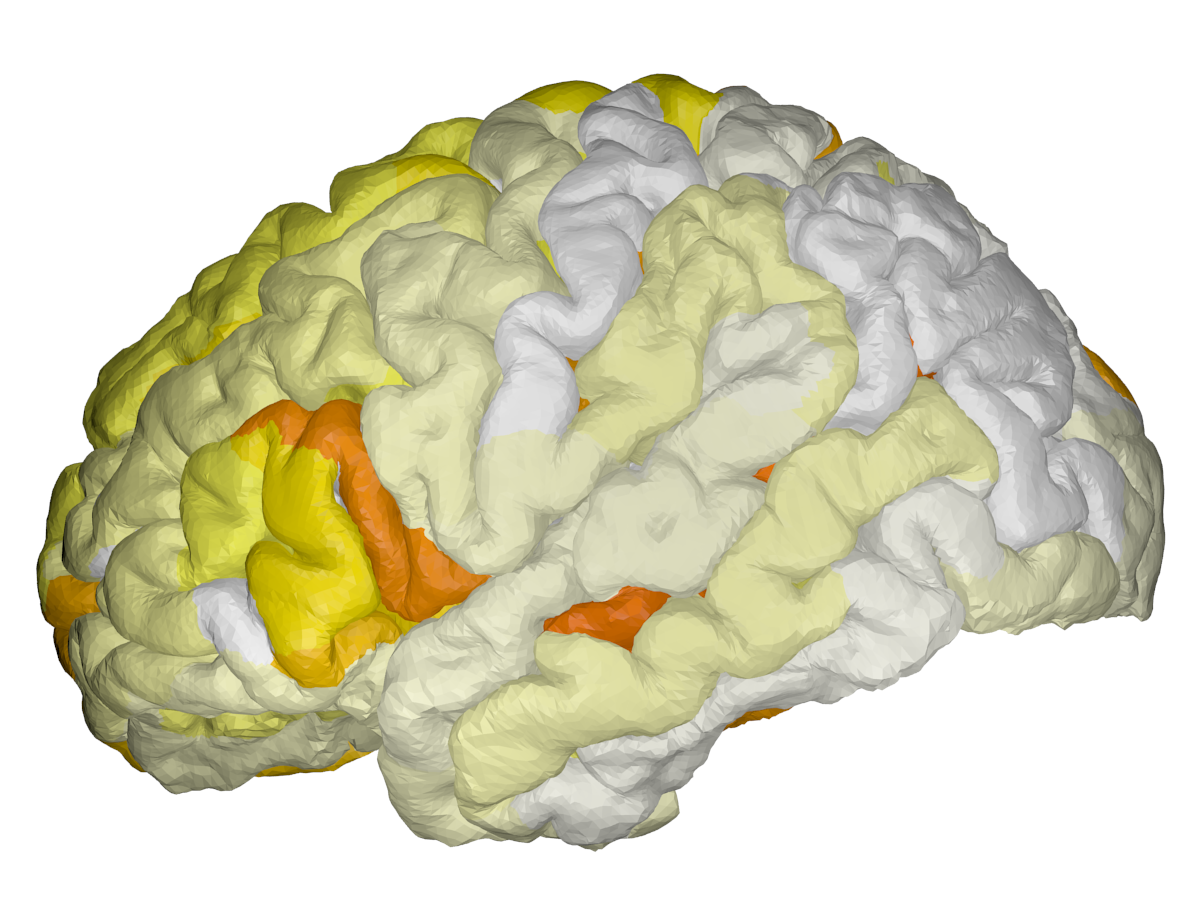}
    & \;\raisebox{0.22\height}[0pt][0pt]{\includegraphics[width=0.05\linewidth]{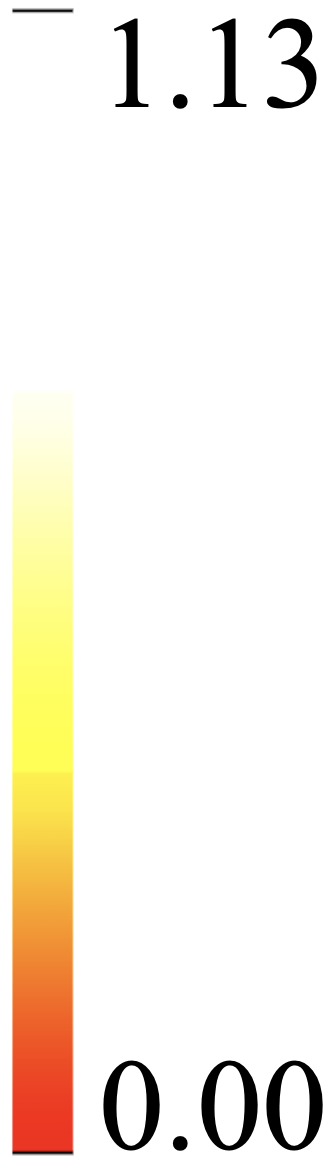}}\\
    \end{tabular}}

    \renewcommand{\arraystretch}{1.0}
    \renewcommand{\tabcolsep}{0.27cm}
    \scalebox{0.7}{
    \begin{tabular}{l|c||l|c||l|c}
        \Xhline{3\arrayrulewidth}
        \multicolumn{2}{c||}{\textbf{Cortical Thickness}}
        & \multicolumn{2}{c||}{\textbf{$\beta$-Amyloid}}
        & \multicolumn{2}{c}{\textbf{FDG}}\\
        \hline
        \textbf{ROI} & \textbf{Scale} & \textbf{ROI} & \textbf{Scale} & \textbf{ROI} & \textbf{Scale}\\
        \hline
         (R) S.cingul.Marginalis & 0.0089 &
         (R) Lat.Fis.ant.Horizont & 0.0662 & 
         (L) sub.putamen & 0.0187 \\
        \hline
         (L) G\&S.occipital.inf & 0.0153 &
         (L) S.oc.middle\&Lunatus & 0.0771 & 
         (L) G\&S.cingul.Mid.Ant & 0.0231 \\
        \hline
         (L) S.front.sup & 0.0252 &
         (R) S.calcarine & 0.0868  & 
         (R) S.temproal.sup & 0.0254 \\
        \hline
         (R) S.suborbital & 0.0254 &
         (R) sub.thalamus & 0.1124  & 
         (R) sub.globus.pallidus & 0.0255  \\
        \hline
         (L) S.oc.temp.med\&Lingual & 0.0387 &
         (R) G.oc.temp.lat.fusifor & 0.1159 & 
         (L) sub.globus.pallidus & 0.0303 \\
        \hline
         (R) S.pericallosal & 0.0387 &
         (L) S.calcarine & 0.1166 & 
         (R) G\&S.cingul.Mid.Ant & 0.0451  \\
        \hline
         (R) S.front.middle & 0.0420 &
         (R) G.temporal.inf & 0.1208 & 
         (L) S.temporal.sup & 0.0609  \\
        \hline
         (R) G.parietal.sup & 0.0500  &
         (R) S.orbital.lateral & 0.1224 & 
         (L) S.calcarine & 0.0662  \\
        \Xhline{3\arrayrulewidth}
    \end{tabular}}
\caption{
Top: 
Visualization of learned scales on the cortical regions of 
left (top) and right (bottom) hemispheres. 
Bottom: 
8 Localized ROIs with the smallest trained scales for classification. (L) and (R) denote left and right hemisphere, respectively.
}
\label{fig:scales}
\end{figure*}

\noindent\textbf{Discussion on the Scales.}
In the pre-clinical AD classification, 
the trained model yields 
node-wise optimized scales, 
where each node corresponds to a specific ROI in the brain.
As the trained scales denote the optimal ranges of ROI-wise neighborhood 
for each modality,
they represent modality-wise characteristics across all ROIs 
providing the interpretability of \ourmodel.
Therefore, 
while ROIs with small trained scales require
information from neighboring ROIs on the classification,
ROIs with large scales need distant features as they are less effective individually.
The learned scales on brain regions
per modality are visualized in 
Fig.~\ref{fig:scales}.
Even for the same region in the brain, 
the local ranges are set differently depending on the modalities, 
which provides multi-dimensional understandings of subnetwork for AD progression.

\begin{figure*}[!t]
\centering
    \renewcommand{\arraystretch}{0.9}
    \renewcommand{\tabcolsep}{0.01cm}
    \scalebox{0.99}{
    \begin{tabular}{ccc}
    \includegraphics[width=0.33\linewidth]{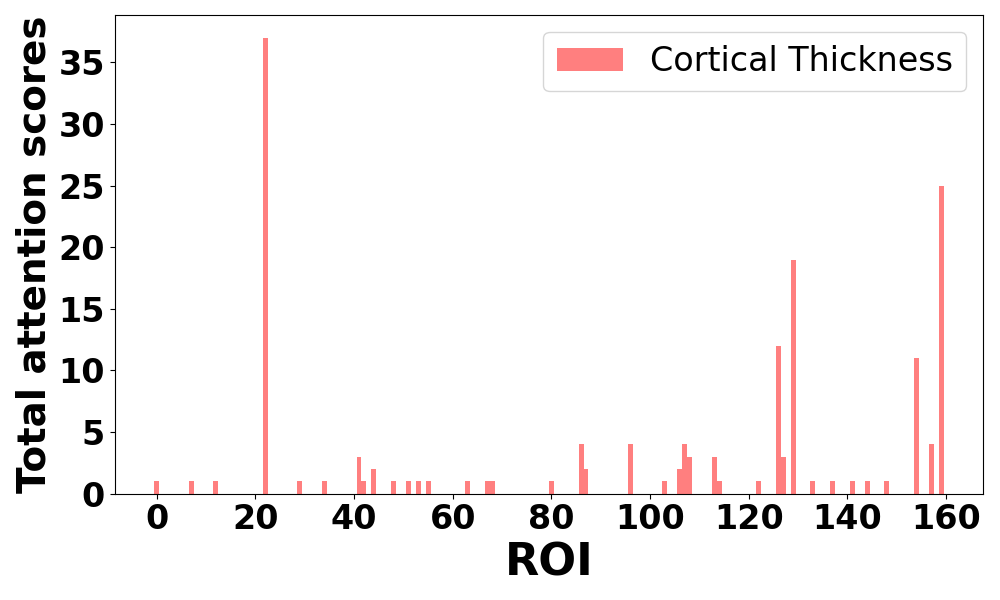}
    & \includegraphics[width=0.33\linewidth]{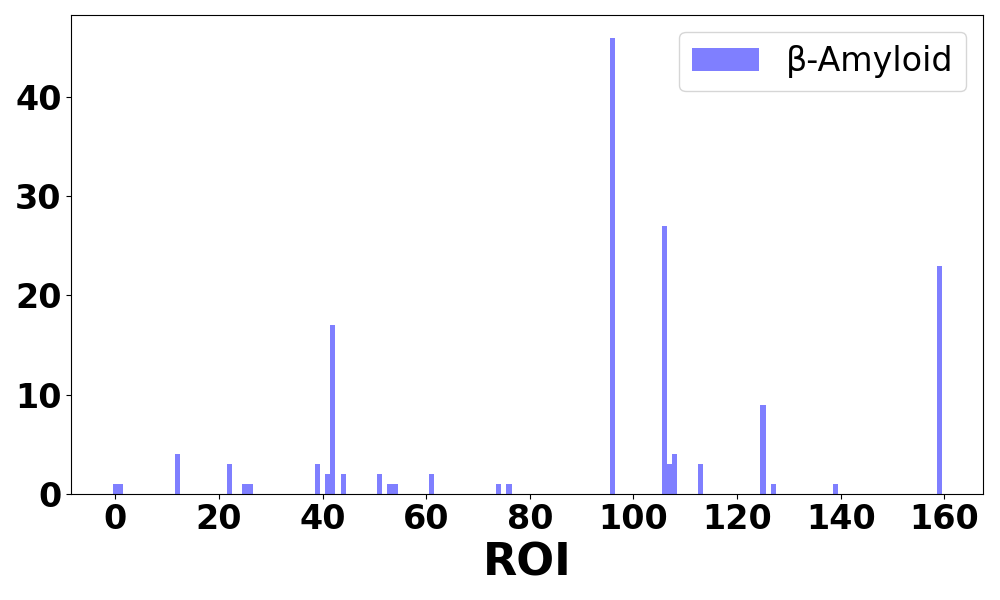}
    & \includegraphics[width=0.335\linewidth]{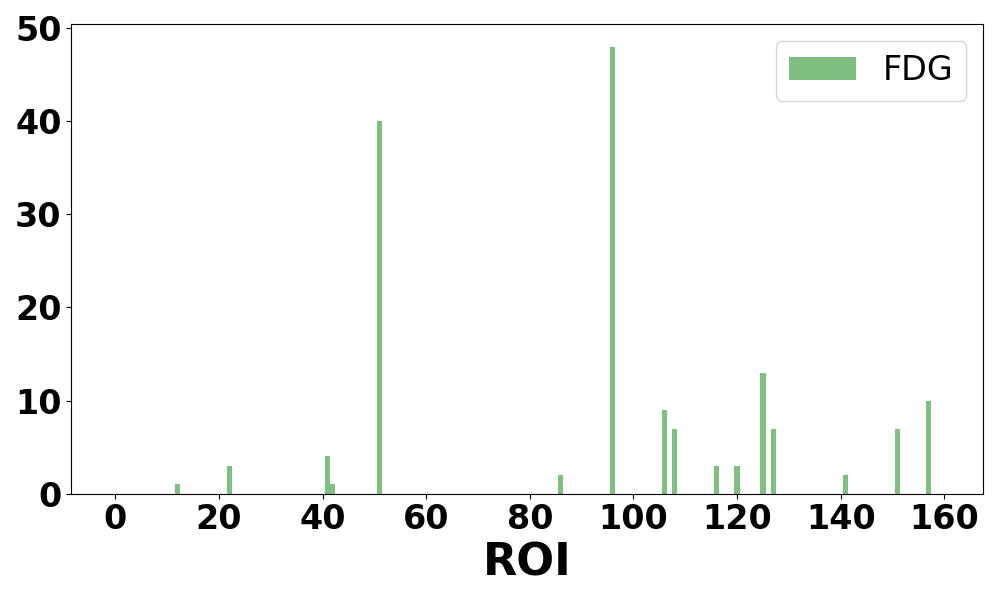}
    \end{tabular}}
    
    \renewcommand{\arraystretch}{1.0}
    \renewcommand{\tabcolsep}{0.16cm}
    \scalebox{0.7}{
    \begin{tabular}{l|c||l|c||l|c}
        \Xhline{3\arrayrulewidth}
        \multicolumn{2}{c||}{\textbf{Cortical Thickness}}
        & \multicolumn{2}{c||}{\textbf{$\beta$-Amyloid}}
        & \multicolumn{2}{c}{\textbf{FDG}}\\
        \hline
        \textbf{ROI} & \textbf{IR} &
        \textbf{ROI} & \textbf{IR} &
        \textbf{ROI} & \textbf{IR} \\
        \hline
        (L) G.oc.temp.med.Lingual & 23.13 \% &
        (R) G.oc.temp.med.Lingual & 28.75 \% &
        (R) G.oc.temp.med.Lingual & 30.00 \%\\
        \hline
        
        (R) sub.putamen & 15.63 \% &
        (R) G.subcallosal & 16.88 \% &
        (L) S.collat.transv.post & 25.00 \%\\
        \hline
        
        (R) S.interm.prim.Jensen & 11.88 \% &
        (R) sub.putamen & 14.38 \% &
        (R) S.collat.transv.post & 08.13 \%\\
        \hline
        
        (R) S.front.inf & 07.50 \% &
        (L) Pole.occipital & 10.63 \% &
        (R) sub.hippocampus & 06.25 \%\\
        \hline
        
        (L) sub.globus.pallidus & 06.88 \% &
        (R) S.collat.transv.post & 05.63 \% &
        (R) G.subcallsoal & 05.63 \%\\
        \Xhline{3\arrayrulewidth}
        \end{tabular}} 
\caption{
Top: 
Distribution of attention scores across all brain regions
with cortical thickness (left), $\beta$-Amyloid (center) and FDG (right).
Bottom: 
Corresponding ROIs with the 5 highest attention scores for classification. 
Importance Rate (IR) indicates how many ROIs pay attention.
(L) and (R) denote left and right hemisphere, respectively.
}
\label{fig:attention}
\end{figure*}

In addition to the visualization of localized scales, 
8 ROIs with the smallest scales for each modality are listed in 
the bottom of 
Fig.~\ref{fig:scales}.
Using the ROI-wise optimized scales with all biomarkers, 
\ourmodels selected most independent ROIs in the subcortical regions (i.e., {\em thalamus, putamen and globus pallidus}), 
temporal regions (i.e., {\em inferior, superior and occipito temporal regions}), 
frontal regions (i.e., {\em middle, superior and orbital
regions}) 
and other important regions that are closely linked to AD.
Based on these results, 
the ROIs with small scales
are significantly important 
in interpreting the classification results depending on the characteristics that each imaging modality captures.  
\noindent\textbf{Pre-clinical AD via ROI Attention.}
From the attention block,
each ROI gains long-range characteristics from other ROIs by modality-wise attention mechanism.
In this regard, 
most relevant ROIs in preclinical AD prediction
can be detected by total attention scores that represent the intensity of attention at each ROI in the brain.
Here, the total attention score is defined as the result of calculating how many ROIs give the highest attention score to the corresponding ROI.
In Fig. \ref{fig:attention}, 
distributions of these scores per ROI 
show which ROIs are making long-range influences. 
Since the distributions of total attention score 
vary across all modalities, 
we can explain which ROI is most important from a specific modality in making predictions.

Top 5 ROIs with the highest importance rate, i.e., the ratio of total attention scores, are listed in the bottom of Fig.~\ref{fig:attention}.
Notably, 
{\em Lingual gyrus} was detected with the highest importance rate from all modalities in common. 
{\em Lingual gyrus}, 
which is especially related to processing logical order of events and encoding visual memories, 
is 
belong to temporal regions and highly linked to AD \cite{lingual1,lingual2}.
In particular, {\em hippocampus} showed a high importance rate in FDG,
and {\em putamen} also simultaneously exhibited a high score in Cortical Thickness and $\beta$-Amyloid.
These regions are one of the first areas to be affected in AD, 
indicating that they are closely associated with pre-clinical AD \cite{hippo,puta}.
From these results, 
we can observe the key regions 
in distinguishing the progressions of neurodegenerative brain diseases 
through modality-wise attentions.

\begin{table}[!t]
    \caption{
    Performance comparisons of different blocks.
    For attention block,
    our multi-modal (MM) attention and existing position-wise attention are compared.
    }
    \centering
    \renewcommand{\arraystretch}{1.0}
    \renewcommand{\tabcolsep}{0.22cm}
    \scalebox{0.8}{
    \begin{tabular}{c|c||ccc}
        \Xhline{3\arrayrulewidth}
        \textbf{Convolution Block} & \textbf{MM Attention} & \textbf{Accuracy} & \textbf{Precision} & \textbf{Recall} \\
        \hline

        \multirow{2}{*}{Multi-Layer Perceptron} 
        & \xmark & 0.939$\pm$0.03 & 0.893$\pm$0.05 & 0.913$\pm$0.04 \\ 
        & \cmark & 0.947$\pm$0.02 & 0.906$\pm$0.04 & 0.933$\pm$0.02 \\ \hline
        \multirow{2}{*}{Graph Convolution Layer} 
        & \xmark & 0.899$\pm$0.01 & 0.835$\pm$0.03 & 0.849$\pm$0.03 \\
        & \cmark & 0.900$\pm$0.01 & 0.834$\pm$0.03 & 0.852$\pm$0.02 \\ \hline
        \multirow{2}{*}{Adaptive Convolution Layer (Ours)} 
        & \xmark & 0.945$\pm$0.03 & 0.903$\pm$0.05 & 0.922$\pm$0.04 \\
        & \cmark & \textbf{0.963$\pm$0.01} & \textbf{0.943$\pm$0.01} & \textbf{0.941$\pm$0.02} \\ \hline

        \Xhline{3\arrayrulewidth}
  \end{tabular}}
\label{tab:ablation}
\end{table}

\noindent\textbf{Ablation Study on the Blocks.}
To explore the effect of each block,
ablation study on convolution types and attention types for preclinical AD classification is given in 
Table~\ref{tab:ablation}.
For the convolution block, Multi-Layer Perceptron (MLP), Graph Convolution and Adaptive Graph Convolution are compared with a choice of multi-head or position-wise attention which was obtained by inputting concatenated features into a single encoder~\cite{transformer}.
The flexible capture of local properties
for each node using adaptive graph convolution 
exhibits better expressive power with 94.5\% accuracy. This metric was boosted up to 96.3\% by the multi-modal attention, demonstrating capturing local and global features with separate blocks but training them jointly is highly effective. 
As the MLP connects all ROIs globally and the Graph Convolution is not adaptively guided by the transformer, the effect of the multi-modal attention was very marginal. 

\section{Conclusion}
\label{sec:conclusion}

In this work, we proposed a novel end-to-end framework \ourmodels to dynamically define node-centric ranges per imaging modality via diffusion kernel,
guided by a subsequent transformer.
Our framework captures local characteristics on graphs by flexibly optimizing node-wise scales separately on imaging modalities,
and obtains a global representation by employing multi-modal self-attention,
which guides the model to better prediction. 
Leveraging multiple imaging measures,
\ourmodels demonstrates superiority
as evidenced by improved performance in preclinical AD classification, and 
the results identifies disease-specific variation through AD-specific key ROIs in the brain. 

\begin{credits}
\subsubsection{\ackname} 
This research was supported by 
NRF-2022R1A2C2092336 (50\%), 
RS-2022-II2202290 (20\%), 
RS-2019-II191906 (AI Graduate Program at POSTECH, 10\%) funded by MSIT,
RS-2022-KH127855 (10\%), 
RS-2022-KH128705 (10\%) funded by MOHW from South Korea.

%

\subsubsection{\discintname}
The authors have no competing interests to declare that are
relevant to the content of this article.
\end{credits}

\bibliographystyle{splncs04}
\bibliography{Paper-1153}

\end{document}